ICDSST 2020 PROCEEDINGS – ONLINE VERSION
THE EWG-DSS 2020 INTERNATIONAL CONFERENCE ON DECISION SUPPORT SYSTEM TECHNOLOGY
I. Linden, M.T. Escobar, A. Turón, F.Dargam, U.Jayawickrama(editors)
Zaragoza, Spain, 27-29 May 2020

# Intelligent Decision Support System for Updating Control Plans

**Fadwa Oukhay**
LISI laboratory, INSAT institute, Carthage University
Tunis, Tunisia
oukhayfadwa@gmail.com

**Pascale Zaraté**
University of Toulouse, IRIT
Toulouse, France
Pascale.Zarate@ut-capitole.fr

**Taieb Ben Romdhane**
LISI laboratory, INSAT institute, Carthage University
Tunis, Tunisia
benromdhane.t@planet.tn

## ABSTRACT

In the current competitive environment, it is crucial for manufacturers to make the best decisions in the shortest time, in order to optimize the efficiency and effectiveness of the manufacturing systems. These decisions reach from the strategic level to tactical and operational production planning and control. In this context, elaborating intelligent decisions support systems (DSS) that are capable of integrating a wide variety of models along with data and knowledge resources has become promising. This paper proposes an intelligent DSS for quality control planning. The DSS is a recommender system (RS) that helps the decision maker to select the best control scenario using two different approaches. The first is a manual choice using a multi-criteria decision making method. The second is an automatic recommendation based on case-based reasoning (CBR) technique. Furthermore, the proposed RS makes it possible to continuously update the control plans in order to be adapted to the actual process quality situation. In so doing, CBR is used for learning the required knowledge in order to improve the decision quality. A numerical application is performed in a real case study in order to illustrate the feasibility and practicability of the proposed DSS.

# INTRODUCTION

Product and process quality control is essential for manufacturing companies in order to achieve their objectives and to meet customer requirements. According to IATF16949 standards [1], automotive manufacturers and their suppliers are required to establish quality control plans for their manufacturing processes. The control plan is a document that defines the quality control actions and methods to be performed for the product characteristics and the process parameters. The aim is to assess the compliance of these characteristics and parameters with the requirements and to minimize their variability. In order to assure efficiency, quality controls need to be focused on the critical features that are important to the customer. In this sense, the control plans optimization is a challenging issue for industrial experts. In fact, the decision concerning what to control, how and where requires considering conflicting criteria. On the one hand, the controls are essential for assuring quality as they prevent the risk of producing and delivering defective products to customers. On the other hand, controlling all products and process characteristics increases strongly the manufacturing costs as well as the time delays.

In order to support the decision process for quality control planning, several research works have been developed. Considerable works are based on multi-objective optimization techniques. These models allow optimizing the control plans based on specific criteria such as total production costs, warranty costs [2], the level of risk and the inspection capacity [3]. Most of these works are subjected to restrictive assumptions and require large volumes of information and precise data. Therefore, they are not adequate to deal with early and complex decision-making issues such as the lack of information availability. To overcome these difficulties, other approaches based on multi-attribute decision-making models are proposed for the development of control plans. An intuitionistic fuzzy decision model is proposed in [4]. It aims to select the best inspection scenario based on expert knowledge and judgment. Three criteria are considered for the decision, namely the cost, the process capability and the external non-conformity rate. Nevertheless, in this work, the criteria are assumed to be independent. This may lead to unsatisfying decision because the criteria may present preferential interactions that should be taken into account [5]. In order to model the criteria interactions, we developed in a previous work a multi-criteria decision making (MCDM) approach for the selection of the best control scenario [6]. This method uses AHP [7] and the Choquet integral [5] for evaluating the performance of the control alternatives according to three main criteria. The latter are the risk priority number, the control cost and the control time. The proposed approach makes it possible to consider the decision makers 'preferences regarding the importance of criteria and their interdependencies.

Indeed, the above-mentioned approaches tackle the decision problem in the quality control planning and permit to generate optimized control plans that are implemented in the manufacturing process. However, these control plans are static and may become ineffective in case the quality circumstances change during manufacturing stage. For this reason, control plans need to be continuously updated in order to be adapted to quality situation changes. For instance, if the process becomes mastered and stabilized, then the controls must be lightened and vice-versa. According to the previous remarks, this paper proposes an intelligent decision support system (DSS) for systematically updating control plans with respect to the actual process quality state. The proposed DSS is a recommender system (RS) that permits to select the best control scenario for the actual situation based on experts' knowledge. Case-based reasoning (CBR) technique is used in order to continuously update the knowledge and improve the decision quality. We present, in the next section, the basic concepts of the DSS, the RS and the CBR. In section 3, the proposed DSS for updating control plans is presented. Sections 4 and 5 are dedicated for the numerical application and the conclusions, respectively.

**BACKGROUNDS**
In this section, we briefly introduce some basic knowledge of the related works.

**Decision support and recommender systems**

Decision Support Systems (DSSs) are popular tools of computerized systems that support the decision-making processes. They are widely used for solving decision problems in different domains e.g. energy, manufacturing, etc. [8]. DSSs have greatly progressed since their appearance. As they are enriched by artificial intelligence techniques, DSSs has evolved from aiding decision makers to perform analysis to providing automated intelligent support. In this sense, Recommender Systems (RSs) are DSSs that are capable of analyzing previous usage behavior and making recommendations for the most suitable items [9]. There are three main types of RS depending on the used approach for the recommendation: The content-based RS, the collaborative RS and the hybrid RS. The content-based RS learns to recommend items that are similar to the ones liked by the user in the past. The collaborative RS makes recommendations to the active user based on items that users having similar tastes liked in the past. The hybrid RS is a combination of the content-based and the collaborative recommendations. Furthermore, several research works used different artificial intelligence techniques for developing personalized RS. We can cite ontology [10] and CBR [11]. In this work, CBR technique is adopted for elaborating the proposed RS.

**Case based reasoning**

CBR is a technique for solving new problems based on specific past experiences that are represented and stored as cases. A case is a set of problems and their associated solutions. A "source case" is a case from which we draw inspiration to solve a new problem called "target case". CBR is a cyclic process that has four main steps: (1) retrieve the most similar cases from databases, (2) reuse the case solutions trying to solve the problem, (3) revise suggested solutions, and (4) retain useful parts of this experience for future problem solving. CBR has been widely used in recommendations. Recently, in [12], CBR is combined with collaborative filtering for elaborating a hybrid RS. The use of CBR approach makes it possible to overcome the cold start problem of the collaborative filtering recommendation and therefore increases the system performance.

In today's highly competitive environment, it is crucial for manufactures to continuously make the best decisions in the shortest time. In this context, developing intelligent techniques for decision-making is a key factor to optimize the efficiency and effectiveness of the manufacturing systems. The decision process in the field of quality control planning is a complex task due to the conflicting criteria and the lack of information availability. The development of an intelligent DSS for quality control planning is then needed in order to enhance the decision-making process.

# PROPOSED DSS FOR UPDATING CONTROL PLANS

In this section, we present the proposed DSS for updating control plans. This DSS is an RS that assists the decision maker (DM) to select the most suitable control scenario for a given quality situation. To do this, CBR technique is used to enable the learning of the required knowledge for the resolution of new cases. The designed methodology is described by the flowchart (Figure 1). We explain the steps of the methodology with respect to the CBR cycle.

**-Case representation:** A case is a piece of knowledge representing an experience and typically a problem and a solution. Cases are usually represented as attribute value pairs that represent the problem and solution features. In this work, the problem represents the quality situation of a given process operation and a given product characteristic. The solution consists of the quality control scenario ($S_i$) that is suitable to deal with this situation while considering the efficiency issue. A quality situation is characterized by four attribute values pairs that represent four process and product quality indicators. The latter are the process capability (Cp), the process capability (Cpk) [13], the internal non-conformity rate (NCR) and the external non-conformity rate (ENCR).

**-Case retrieval:** This phase consists in searching in the cases base for a similar case (source case) that will be used to solve the new case (target case). In this work, similarity between the target case ($C^t$) and the source case ($C^s$) denoted $Sim(C^t, C^s)$, is calculated using the Minkowski distance as follows: $Sim(C^t, C^s) = |Cp^t - Cp^s| + |Cpk^t - Cpk^s| + |NCR^t - NCR^s| + |ENCR^t - ENCR^s|$. The source cases stored in the cases base are prioritized according to their similarities. The DM defines a threshold. Thus, the source case that has the smallest distance inferior to the threshold is selected to be used. However, the cases base may be initially empty or similar cases may not be found. Therefore, the solution for the target case is chosen manually by the DM according to his knowledge and preferences. In this case, the proposed MCDM method in our previous work [6] is used for the selection of the best control scenario.

**-Case adaptation:** The phase of adaptation in the CBR cycle is the process of proposing a solution to a new problem from solutions belonging to the recalled source cases. In this work, the system automatically recommends the solution of the source case to solve the target case. However, the DM has to validate the recommended control scenario to be used. If the he estimates that the recommended solution is not suitable, then the manual option is used.

**-Case revision:** During the revision phase, the selected solution is evaluated. For doing so, the DM defines the objectives ($Cp^*$, $Cpk^*$, $NCR^*$ and $ENCR^*$) to be reached using the proposed solution. The latter is applied in the manufacturing process. After a T period defined by the DM, the obtained results are compared with objectives. If objectives are reached, the solution is then judged satisfactory and therefore the case (problem-solution) is considered relevant. However, if the results are not acceptable, then the case have to be repaired. In this work, two types of revision are performed: The revision of the manual choice and the revision of the automatic choice. In case the solution is manually chosen, then the DM revises his evaluations in the decision-making process. In fact, the pairwise comparisons in the AHP matrix are adjusted. However, in case the solution is selected automatically by the RS and if the obtained results of Cp, Cpk, NCR and ENCR are not satisfactory, then the case is not sufficiently similar. Therefore, the similarity threshold should be adjusted.

**-Case retaining**: The retaining or learning phase consists in incorporating what is useful in the cases base and synthesizing the new knowledge that will be reused later. Therefore, the storage of new relevant cases enriches the cases base and increases the system experience.

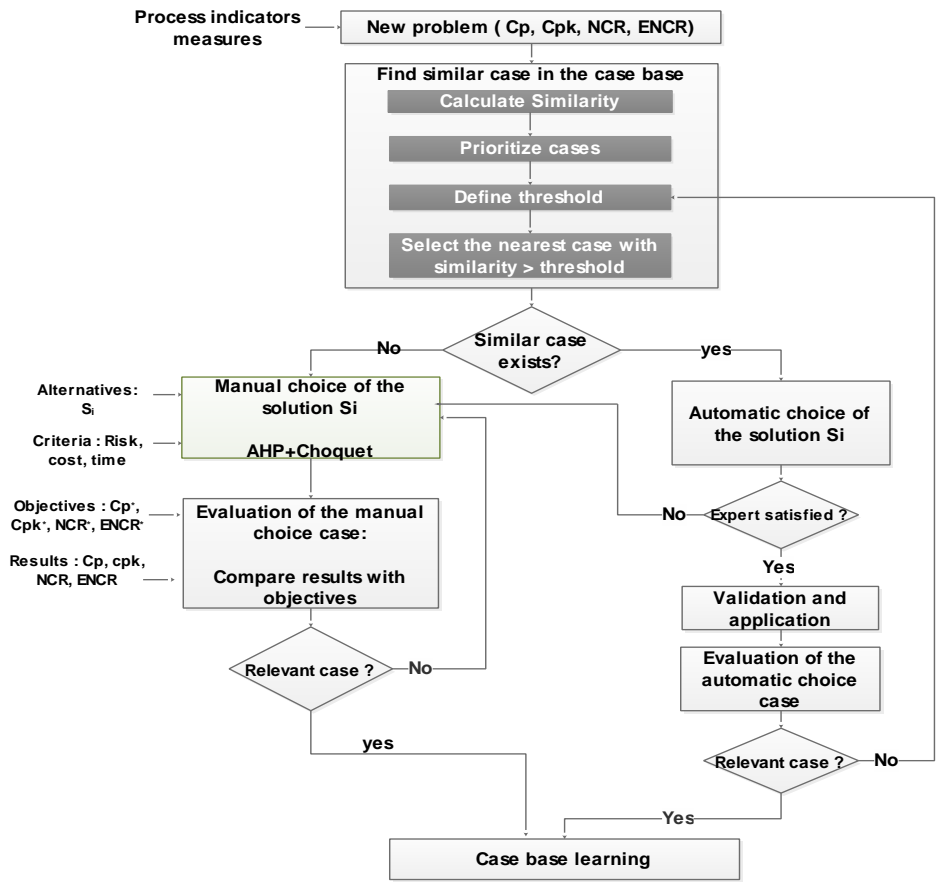

Figure 1: Flowchart of the proposed methodology

## NUMERICAL APPLICATION

In order to illustrate the proposed methodology, we perform a numerical application in a real case study. The latter is an enterprise of automobile components, which manufactures different wires to command airbags [6]. We present, in what follows, the decision process for the quality control plan selection concerning a process operation (Splitting/Crimping) and a product characteristic (crimping height). The problem attributes are entered by the DM in the DSS (Figure2). Given this quality situation, the DM performs the manual evaluation of the control scenarios according to the considered criteria. The evaluations of the alternatives and the selected control scenario are presented in Figure 3. According to the manual evaluations performed by the DM using AHP method and Choquet integral, the selected control scenario is S2 "Sampling control by measure (simple plan)". Therefore, the DM agrees the choice and S2 is applied for the considered process operation. Otherwise, the DM repeats the manual evaluations. In order to illustrate the automatic choice, some relevant cases are added to the cases base. The DM enters new target case attributes. The system finds similar case in the cases base. Thus, the solution S3 is recommended by the system for solving the target case. At this step, the DM can display the details of the source case and agree the recommendation or disagree and perform the manual choice (see Figure 4).

Figure 2: entering the quality situation in the DSS

Figure 3: The obtained alternatives evaluation using the manual choice option.

Figure 4: The automatic choice illustration.

**CONCLUSIONS**

This paper presents an intelligent DSS for updating quality control plans. The proposed DSS is an RS that makes it possible to select the best control scenario for a given quality situation using two different approaches. The first is a manual choice based on an MCDM method. The other is an automatic recommendation using CBR. The effectiveness of the DSS to improve the decision process is illustrated by its application on a real case study. In future work, we are interested in optimizing the process of calculating similarities in order to improve the accuracy of the decision.